\documentclass[runningheads]{llncs}
\usepackage{amssymb}
\DeclareUnicodeCharacter{25D1}{\textcircled{\textbf{\textminus}}}
\usepackage{newunicodechar}
\newunicodechar{✘}{\texttimes}
\usepackage[utf8]{inputenc}
\DeclareUnicodeCharacter{2714}{\checkmark}
\usepackage{pifont}

\usepackage[utf8]{inputenc}
\usepackage{amsmath}
\usepackage{esvect}

\usepackage{caption}
\usepackage{amsmath}
\usepackage{array}     
\usepackage{booktabs}
\usepackage[table]{xcolor}    
\usepackage{makecell}

\usepackage[T1]{fontenc}
\usepackage{graphicx}
\usepackage{hyperref}
\usepackage{color}

\begin{document}
\title{Pretraining Data Exposure in Large Language Models: A Survey of Membership Inference, Data Contamination, and Security Implications}
%
%
\author{Ziyi Tong\orcidID{0009-0007-5701-9705} \and
Feifei Sun\orcidID{0009-0008-7001-9977} \\
\and
Le Minh Nguyen\orcidID{0000-0002-2265-1010}}
\authorrunning{Z. Tong et al.}
%
\institute{Japan Advanced Institute of Science and Technology, Japan \\
\email{s2320416@jaist.ac.jp, nguyenml@jaist.ac.jp}}
\maketitle              
\begin{abstract}
Large Language Models (LLMs) have become the predominant paradigm in NLP, advancing both research and industry. As model sizes and pretraining data grow, concerns about Pretraining Data Exposure (PDE) increase due to the scale and opacity of training datasets.

PDE refers to determining whether specific data appeared in an LLM's pretraining corpus. It is critical for ensuring evaluation integrity and protecting privacy, intersecting two key areas: data contamination and membership inference. Though conceptually related, these areas have often been studied in isolation. This paper offers the first unified survey of both under the PDE framework. We formalize PDE across exposure levels, review attack and defense methods, synthesize empirical findings, and highlight open challenges and future research directions.

\end{abstract}
\section{Introduction}

The rapid development of Large Language Models (LLMs)\cite{anthropic2024claude3,team2023gemini} along with the increasing size of pretraining datasets has introduced new challenges: how can we determine whether specific data was included in an LLM's pretraining corpus?

LLMs are trained on largely opaque datasets\cite{shi_detecting_2024}, often collected via automated web crawlers, making it impossible to determine which data points were included in the training. This raises serious security and privacy concerns \cite{Balloccu_2024,ishihara2023training,cheng_survey_2025,sainz_nlp_2023,jacovi_stop_2023,Mireshghallah_2022}. Public evaluation datasets are particularly vulnerable to contamination\cite{oren_proving_2023,xu2024benchmark}, test-train overlap compromises validity, as models may memorize rather than generalize. 

These concerns make it imperative to study Pretraining Data Exposure (PDE)—the problem of verifying whether specific data has been included in an LLM's pretraining corpus. In this paper, we approach PDE from a unified perspective by bridging two existing research domains: (1) data contamination\cite{cheng_survey_2025,fu2024does}, which investigates overlaps between training and evaluation datasets, and (2) membership inference\cite{hu2022membership,Shokri_2016}, which seeks to determine whether specific instances were part of the pretraining data. While MIA is often mentioned in data contamination studies, previous research\cite{cheng_survey_2025,hu2022membership,deng2024unveiling,fu2024does,ishihara2023training} has typically treated it as a minor subset of the broader contamination problem. We position both fields as equally important, systematically integrating their latest attack and defense strategies into a comprehensive literature review. 

At their core, both data contamination and membership inference share a common objective: \textit{Determining whether a specific data point exists within the pretraining corpus of an LLM.} 

Specifically, our key contributions are as follows.

1. \textbf{A unified review of PDE across two domains.} We propose a comprehensive framework for PDE, encompassing both instance-level and dataset-level exposure. To our knowledge, this is the first work to treat membership inference as equally important as data contamination, rather than as a subset. While the two areas overlap, we argue that MIA offers additional insight by enabling instance-level PDE detection.

2. \textbf{A novel taxonomy of PDE attack and defense strategies.} We present a new taxonomy for PDE attacks and defenses, organized by real-world scenarios and user types, to better align existing research with practical deployment contexts.

3. \textbf{Latest updates and future directions.} We provide an up-to-date review of PDE research, highlight key challenges and open problems, and propose directions for future work.


\section{Background}
In this section, we introduce key terms and concepts to provide a clear foundation for understanding the Pretraining Data Exposure (PDE) problem.  

\textbf{Membership Inference} 
Membership inference is an attack technique in which an adversary seeks to determine whether a specific data sample was part of a machine learning model's training dataset\cite{Mozaffari_2024,Duan_2024,hu2022membership}. In the context of LLMs, membership inference attacks (MIAs) leverage the model's outputs\cite{Duan_2024}, probability distributions\cite{6bf34b4a1937ca5ae692594eda880ff671b8ee57,shi_detecting_2024}, gradients\cite{Mancera_2025,d27fd3c0e09539db1e5251a3ba6a2d8f140ff546}, or internal representations to differentiate between member instances (samples seen during training) and non-member instances (unseen data). These attacks pose significant privacy risks, especially when models inadvertently memorize and disclose sensitive or private information\cite{d1101476c85ae324142440e9f568ecbf41625be5}.

\textbf{Data Contamination}
Data contamination is universally defined as the inclusion of evaluation data (input and/or labels) in training datasets of models \cite{deng2024unveiling,cheng_survey_2025,sainz_nlp_2023,Palavalli_2024}. Contamination causes models to memorize instead of generalize, inflating benchmark performance and undermining evaluation validity and generalization. Larger models often exhibit stronger inflation due to memorization\cite{Liu_2023}. Traditional approaches for contamination detection include N-gram overlap, exact-match comparisons\cite{Palavalli_2024}, and perplexity analysis\cite{li2023estimatingcontaminationperplexityquantifying}, but these methods falter with approximate, noisy, or adversarial contamination\cite{samuel_towards_2024,Dekoninck_2024}. Conventionally, strategies for preventing and mitigating data contamination focus on dataset design and curation \cite{Zhu_2024}, as well as innovations in evaluation protocols\cite{sainz_nlp_2023}.

\section{Definitions}
This section formalizes PDE with mathematical definitions at both instance and dataset levels. Instance-level PDE aligns with membership inference attacks (MIA), while dataset-level PDE corresponds to data contamination. Therefore, following previous work\cite{fu2024does}, we provide definitions of instance-level PDE and dataset-level PDE.
\subsection{Instance-Level PDE }

Let \( D_M \) denote the pretraining data of an LLM \( M \). The binary function \( f(M, x) \) determines whether an individual instance \( x \) is seen by the model \( M \):

\begin{equation}
f(M, x) =
\begin{cases} 
1 & \text{if } \exists x' \in D_M, \, b(x, x') = 1 \\ 
0 & \text{if } \forall x' \in D_M, \, b(x, x') = 0
\end{cases}
\end{equation}

If \( f(M, x) = 1 \), the instance \( x \) is considered \textbf{exposed} (seen by the model).
If \( f(M, x) = 0 \), the instance \( x \) is considered \textbf{unexposed} (unseen by the model).

\subsection{Dataset-Level PDE}

A dataset \( D \) is \textbf{exposed} (partially seen by \( M \)) if at least one instance \( x \) in \( D \) is seen:

\begin{equation}
\exists x \in D, f(M, x) = 1
\end{equation}

A dataset is \textbf{fully exposed} if all instances within \( D \) are seen:

\begin{equation}
\forall x \in D, f(M, x) = 1
\end{equation}

\subsection{Exposure Score for PDE}
In addition to the binary definition, we define an exposure score to quantify how much of \( D \) has been seen by \( M \):

\begin{equation}
PDE(D, M) = \frac{\sum_{x \in D} f(M, x)}{|D|}
\end{equation}

where \( PDE(D, M) \) represents the proportion of exposed instances in dataset \( D \).

If \( PDE(D, M) = 1 \), the dataset is fully exposed.
If \( PDE(D, M) = 0 \), the dataset is fully unexposed.
If \( 0 < PDE(D, M) < 1 \), the dataset is partially exposed.

\section{Threat Models and Detection Strategies for PDE}
\subsection{Threat Model}
We define the threat model for text-based LLMs, considering adversaries with query access but no access to pretraining data. Our analysis focuses on English-language models, including both open-source models and commercial APIs.

\subsection{Taxonomies of Detection strategies}
Unlike previous studies\cite{cheng_survey_2025,hu2022membership,deng2024unveiling,fu2024does,ishihara2023training}, we categorize PDE detection methods based on real-world application scenarios and user types. Our taxonomy aligns existing research more closely with practical deployment settings. Empirically, we identify four key LLM application scenarios where pretraining data exposure presents significant challenges. \autoref{tab:security_risks} presents a taxonomy of scenarios, detailing descriptions, user types, and related security risks, offering a structured view of stakeholder interactions and vulnerabilities.

Besides the papers mentioned in \autoref{tab:security_risks}, \textbf{benchmark contamination} scenario was also investigated in studies: \cite{li2023estimatingcontaminationperplexityquantifying,Dekoninck_2024,1ea243f1b697aae22e6f0349fa64857780a6108a,84725855d10b531eb8cbe54935dda0440c2fc750,li_task_2023}. Approaches such as perplexity analysis\cite{li2023estimatingcontaminationperplexityquantifying}, exchangeability tests\cite{oren_proving_2023}, and statistical tools like ConStat\cite{54311e1b1250e87194bcaf4036bfefd8d5ce5acf} offer mechanisms for identifying memorization and dataset overlap. However, detecting paraphrased or partially contaminated instances remains challenging, especially in black-box settings.

In \textbf{personal data exposure} scenario, beyond the works listed in \autoref{tab:security_risks}, these studies also explored the personal data exposure problem: \cite{carlini_quantifying_2023,Vakili_2023,Wen_2024}. Overall, techniques like differential privacy reduce utility but fail to reliably protect rare or sensitive PII \cite{d1101476c85ae324142440e9f568ecbf41625be5}. Innovations like MemHunter allow dataset-wide PII leakage verification with reduced computational costs\cite{df9f16443980cb9f0dfdd3c492c9de887b71a4eb}. However, detecting and addressing PII leakage remains a critical goal.

\begin{table}[h]
    \caption{Taxonomies of PDE detection methods based on LLM application scenarios and user types, associated with security risks.}
    \label{tab:security_risks}  
    \centering
    \setlength{\tabcolsep}{6pt}
    \renewcommand{\arraystretch}{2.1}  
    \large  
    \resizebox{\textwidth}{!}{  
    \rowcolors{2}{gray!15}{white} 
    \begin{tabular}{|p{4.3cm}|p{1.8cm}|p{5.8cm}|p{5.8cm}|p{4.2cm}|}
    \toprule
    \textbf{Application Scenario} & \textbf{User Types}& \textbf{Scenario Description} & \textbf{Security Risks}& \textbf{Relevant Researches} \\ \midrule
    NLP Benchmark Contamination & LLM developer & LLMs are trained on publicly available NLP benchmarks, leading to inflated evaluation results because the model has already seen the test set. & Model performance evaluation becomes unreliable, making it unclear whether the model is truly generalizing or just memorizing test examples. & e.g.\cite{deng_investigating_2024,Yang_2023,Zhang_2024,Chen_2025,ravaut_how_2024,Balloccu_2024,oren_proving_2023}\\

    Personal Data Exposure from Web Crawling &API User& LLMs scrapes massive web corpora, accidentally ingesting personal information, social media posts, and leaked databases. & If models memorize personal data, they might regurgitate sensitive details when prompted. & e.g.\cite{d1101476c85ae324142440e9f568ecbf41625be5,117ede00f84edce011d2d8a9142f3877a24c2842,Wei_2024}\\

    Copyrighted Content \& Intellectual Property Risks &API Provider& LLMs trained on web data ingest copyrighted content, leading to legal concerns. & LLMs may directly output copyrighted content, raising ethical and legal issues. & e.g.\cite{shi_detecting_2024,55a9e7e09d3ff5df147cf4ed85f0387a4d5da149,cde8783eebe3cfafef36edf22bdc2f40cbd3e0d4,e472563c4816a2c70aab0652c184087f3cfd2bda,68bb0cbe1d9d209fcbef96949da728f42063b73a}\\

    Code \& Software Security Risks &API Provider and API User& LLMs trained on public code repositories (e.g., GitHub, Stack Overflow) may leak proprietary or insecure code. & Models output license-violating code or insecure snippets. & e.g.\cite{7d75b26b835292750aa199230c4a88ffee339a28,2f2eb528d5a5415abf760b779e7108aa24116985,266d35fa042220236d25b8f7101914a44df4febd,Yang_2023,0a5ca7649ab378cc8c734ddac5bf6c6c00f086c1}\\ 
    
    \bottomrule
    \end{tabular}
    }
\end{table}

In the \textbf{copyrighted content} scenario, Expectation-Maximization MIA (EM-MIA): A state-of-the-art inference method refines membership probabilities, excelling in detecting copyrighted data under experimental setups\cite{68bb0cbe1d9d209fcbef96949da728f42063b73a}. Sampling-based MIAs (SaMIA): Improves inference without internal training dataset access\cite{33b81b3b25c84674936e83bf91bef7d5af870ee2}. Overall, despite progress in addressing copyright contamination and MIA risks, no single solution fully resolves the technical, legal, and ethical challenges, highlighting the need for integrated detection strategies.

In the \textbf{code \& software security risks} scenario, papers mentioned in \autoref{tab:security_risks}, confirm significant risks of PDE targeting programming-related text in LLMs, providing tailored methods (e.g., CodeMI\cite{2f2eb528d5a5415abf760b779e7108aa24116985}, TraWiC\cite{266d35fa042220236d25b8f7101914a44df4febd}), metrics (e.g., token-level decoding\cite{5e8b373f921977f15844cad8c24bbeb1cd6484e6}, confidence score calibration\cite{7d75b26b835292750aa199230c4a88ffee339a28}). However, programming-related text poses elevated risks due to deterministic structures, leading to the memorization of sensitive proprietary content (e.g., credentials, function templates). As a result, a high false positive rate arises in membership inference metrics. Moreover, deduplication strategies struggle with semantic equivalence in programming corpora\cite{deng_investigating_2024}, limiting their reliability in detecting nuanced contamination scenarios.

\section{Mitigation and Defense Mechanisms for PDE}
This section outlines contamination mitigation and membership inference defenses.

\textbf{Dynamic benchmark.} A dynamic benchmark is a regularly updating test dataset. Dynamic benchmarking offers a promising approach to mitigating data contamination in model evaluation \cite{Fan_2025,Qian_2024,Li_2023,Zhu_2024,Chen_2025,White_2024}. For the coding area, there is \cite{afe0998d191f3ea8490c7df100a3ffc5dcc62c5e}. Proposed evaluation methods, such as \textbf{contamination-free benchmarks} prevent test-train overlap and \textbf{dynamic dataset splits} update test sets post-training to ensure exclusion from pretraining data. These methods improve benchmark robustness by emphasizing generalization over memorization in the context of rapidly evolving LLMs.\cite{cheng_survey_2025,Cao_2024}.

\textbf{Private and Secure Benchmarking.} A private, secure benchmark prevents data contamination by preserving dataset integrity and confidentiality during evaluation. \cite{jacovi_stop_2023} underscores the importance of protecting future datasets and provides practical strategies, including the encryption of test datasets. Cryptographic isolation and confidential computing frameworks were proposed to secure test datasets from contamination\cite{7274790327898d95d807c856caa5aeba7270f3c1}. A secure benchmark prevents unintentional contamination and enables more reliable and trustworthy evaluation.

\textbf{Automated Decontamination.} Systems like AntiLeak-Bench\cite{e8b3f92bd5b09ee90f74d3b82b60b3c2c796df33} use automated frameworks to create contamination-free benchmarks. \cite{Li_2024} presents a systematic strategy for preventing contamination, which includes periodic crawlers and a contamination detection mechanism. \cite{Zhu_2024} tackles the contamination problem by identifying and modifying leaked samples, ensuring that their difficulty level remains unchanged.

\textbf{Watermarking}. \cite{fd4ccf695be157c925a62c163d7552124721e501} demonstrates that watermarking is effective for copyright protection and reduces the success rate of model inversion attacks. \cite{78c516bc91be667fa25115c9d5c029ec3ac210da} brings TextMarker, which employs a backdoor-based membership inference method to protect sensitive data in pre-trained models, though it increases training complexity. Watermarking supports intellectual property enforcement, but its effects on training complexity and model robustness remain underexplored.

\textbf{Machine unlearning.} Machine unlearning is, ideally, the ultimate solution for removing data from pre-trained models. Efforts have been made to leverage token-specific characteristics\cite{Tran_2025}, benchmark the real-world knowledge unlearning \cite{Jin_2024}, advocate for a minority-aware evaluation framework\cite{Wei_2024}, and erase famous characters from the LLM\cite{eldan2023s,liu2024revisiting}. However, the effectiveness of unlearning remains limited and requires further refinement.

Defending against Pretraining Data Exposure (PDE) involves trade-offs among privacy, robustness, scalability, and transparency. Preventative methods like dynamic and secure benchmarks reduce contamination but may limit reproducibility and openness. Automated decontamination scales well but struggles with paraphrased content. Watermarking aids intellectual property enforcement but can affect model generalization and remains legally uncertain. Machine unlearning enables post hoc data removal for privacy compliance, but is technically challenging and may impact performance. No single method suffices; effective PDE defense requires combining strategies tailored to deployment needs and data sensitivity.

\subsection{Comparison of SOTA PDE Methods}
As shown in \autoref{tab:method_comparison}, we provide a comparison of SOTA PDE methods based on code availability,
dataset availability and benchmark availability.

\begin{table*}[t]
    \centering
    \setlength{\tabcolsep}{4pt}
    
    \footnotesize 
    \caption{Comparison of SOTA PDE Methods Based on Code Availability, Dataset Availability, Benchmark Availability.}
    \label{tab:method_comparison}
    \small
    \renewcommand{\arraystretch}{1} 
    \resizebox{\textwidth}{!}{  
    \begin{tabular}{lcccccc}
        \toprule
        \textbf{Method} & \textbf{Level}& \textbf{Types} & \textbf{Code Available?} & \textbf{Dataset Info?} & \textbf{Benchmark Info?}  \\
        \midrule
        \cite{Balloccu_2024} & Dataset & Detection & \textcolor{red}{✘} & \textcolor{red}{✘} & \textcolor{green}{\checkmark} \\
        \cite{68bb0cbe1d9d209fcbef96949da728f42063b73a} & Instance & Detection &\textcolor{green}{\checkmark}&\textcolor{green}{\checkmark}&\textcolor{green}{\checkmark}\\
        \cite{oren_proving_2023}&Dataset&Detection&\textcolor{green}{\checkmark}&\textcolor{green}{\checkmark}&\textcolor{green}{\checkmark}\\
        \cite{shi_detecting_2024}& Instance& Detection&\textcolor{green}{\checkmark}&\textcolor{green}{\checkmark}&\textcolor{green}{\checkmark}\\
        \cite{deng_investigating_2024}&Dataset&Detection&\textcolor{red}{✘}&\textcolor{green}{\checkmark}&\textcolor{green}{\checkmark}\\
        \cite{Mattern_2023}&Instance&Detection&\textcolor{green}{\checkmark}&\textcolor{green}{\checkmark}&\textcolor{red}{✘}\\
        \cite{Li_2023}&Dataset& Mitigation&\textcolor{green}{\checkmark}&\textcolor{green}{\checkmark}&\textcolor{green}{\checkmark}\\
        \cite{e8b3f92bd5b09ee90f74d3b82b60b3c2c796df33}&Dataset&Mitigation&\textcolor{green}{\checkmark}&\textcolor{green}{\checkmark}&\textcolor{green}{\checkmark}\\
        \cite{eldan2023s}&Instance&Mitigation&\textcolor{red}{✘}&\textcolor{green}{\checkmark}&\textcolor{green}{\checkmark}\\

        \bottomrule
    \end{tabular}
    }
\end{table*}

\section{Challenges and Future Directions}
 Despite progress in detecting PDE, major challenges remain. Identifying paraphrased, partially contaminated instances or low-occurrence instances is especially difficult. No single approach fully addresses the technical, legal, and ethical issues of PDE, underscoring the need for integrated solutions. Programming-related text poses added risk due to its deterministic nature, making it more prone to memorization and leakage of sensitive or proprietary content.  

To address these challenges, future research should explore several promising directions. One is developing \textbf{unlearning} techniques that enable models to forget specific data points post-training without compromising overall performance. Advancements in \textbf{model explainability} can provide deeper insights into how models memorize and reproduce training data, ultimately leading to more effective mitigation strategies. \textbf{Semantic-level detection} enables identifying contamination beyond exact matches by capturing meaning-based similarities instead of relying solely on lexical overlap. Advancing this area can promote more secure, transparent, and accountable handling of PDE in LLMs.

\section{Conclusion}
This paper presents a comprehensive review of pretraining data exposure (PDE), unifying research on instance-level extraction from membership inference attacks and dataset-level extraction from data contamination. We summarize detection and defense methods, categorize them by user types and application scenarios, and analyze the strengths and limitations of the current work. Finally, we outline future directions to support deeper understanding and continued progress in this rapidly evolving field.

\section{limitation}
Despite our efforts to be comprehensive, some relevant studies may have been omitted. This review focuses solely on text-based Large Language Models (LLMs), excluding other model types, multimodal systems, and research in multilingual or low-resource settings.

\bibliographystyle{splncs04}
\bibliography{reference}
\end{document}